\documentclass{article}
\usepackage{spconf,amsmath,graphicx}
\usepackage{booktabs}
\usepackage{amsmath}
\usepackage{hyperref}
\usepackage{amssymb}
\usepackage{enumitem}
\usepackage{xcolor}
\usepackage{color, colortbl}
\usepackage{tabularx}

\title{The complementary roles of non-verbal cues for \\Robust  Pronunciation Assessment}

\name{Yassine El Kheir, Shammur Absar Chowdhury, Ahmed Ali}
\address{Qatar Computing Research Institute, HBKU, Doha, Qatar
}
\begin{document}
%
\maketitle
\begin{abstract}


Research on pronunciation assessment systems focuses on utilizing phonetic and phonological aspects of non-native (L2) speech, often neglecting the rich layer of information hidden within the non-verbal cues. In this study, we proposed a novel pronunciation assessment framework, \textbf{IntraVerbalPA}.
The framework innovatively incorporates both fine-grained frame- and abstract utterance-level non-verbal cues, alongside the conventional speech and phoneme representations. Additionally, we introduce ``Goodness of phonemic-duration'' metric to effectively model duration distribution within the framework. Our results validate the effectiveness of the proposed IntraVerbalPA framework and its individual components, yielding performance that either matches or outperforms existing research works.

\end{abstract}
\section{Introduction} \label{sec:intro}
Computer-assisted pronunciation training (CAPT) for foreign language learning has seen a surge in global demand in recent years.
CAPT benefits non-native learners with personalized, cost-effective feedback, promotes self-directed learning and improves pronunciation skills. It also offers flexibility compared to traditional instruction \cite{eskenazi2009overview,litman2018speech}.
One of the main objective of the CAPT is to automate pronunciation assessment (PA).
To achieve this goal, the automated PA model need to estimate a score that reflects the oral proficiency based on some standardized assessment criteria \cite{levy2013call, eskenazi2009overview}. 

The task of PA is inherently subjective, even scores assigned by human expert annotator often vary for the same spoken utterance. These discrepancies arises from annotator's unique experiences, their own interpretations of the scoring guidelines, and/or their focus on specific aspect of pronunciation -- like fluency, prosody, word accuracy or even a combination.
Hence, designing an automated PA that emulate the annotators' (or a teacher) is very much complex and challenging. The challenges extends beyond the constraints of datasets availability, and modeling intricacies, to include the crucial task of selecting features and approach to model their representations.

Numerous investigations have explored a range of features and modeling approaches aimed at enhancing modeling performance. These explorations have encompassed the utilization of Goodness-of-Pronunciation (GOP) metrics \cite{lin20f_interspeech,JIM, hu15_slate}, the integration of manually crafted handful of non-verbal features such as duration, energy, and pitch \cite{zhang2021multilingual,3M,chen2023multipa}, as well as the utilization of state-of-the-art pre-trained self-supervised learning models for modeling improvement \cite{kim2022automatic,10022486,yang2022improving}. 




\begin{figure*} 
\centering
\includegraphics[width=0.7\textwidth]{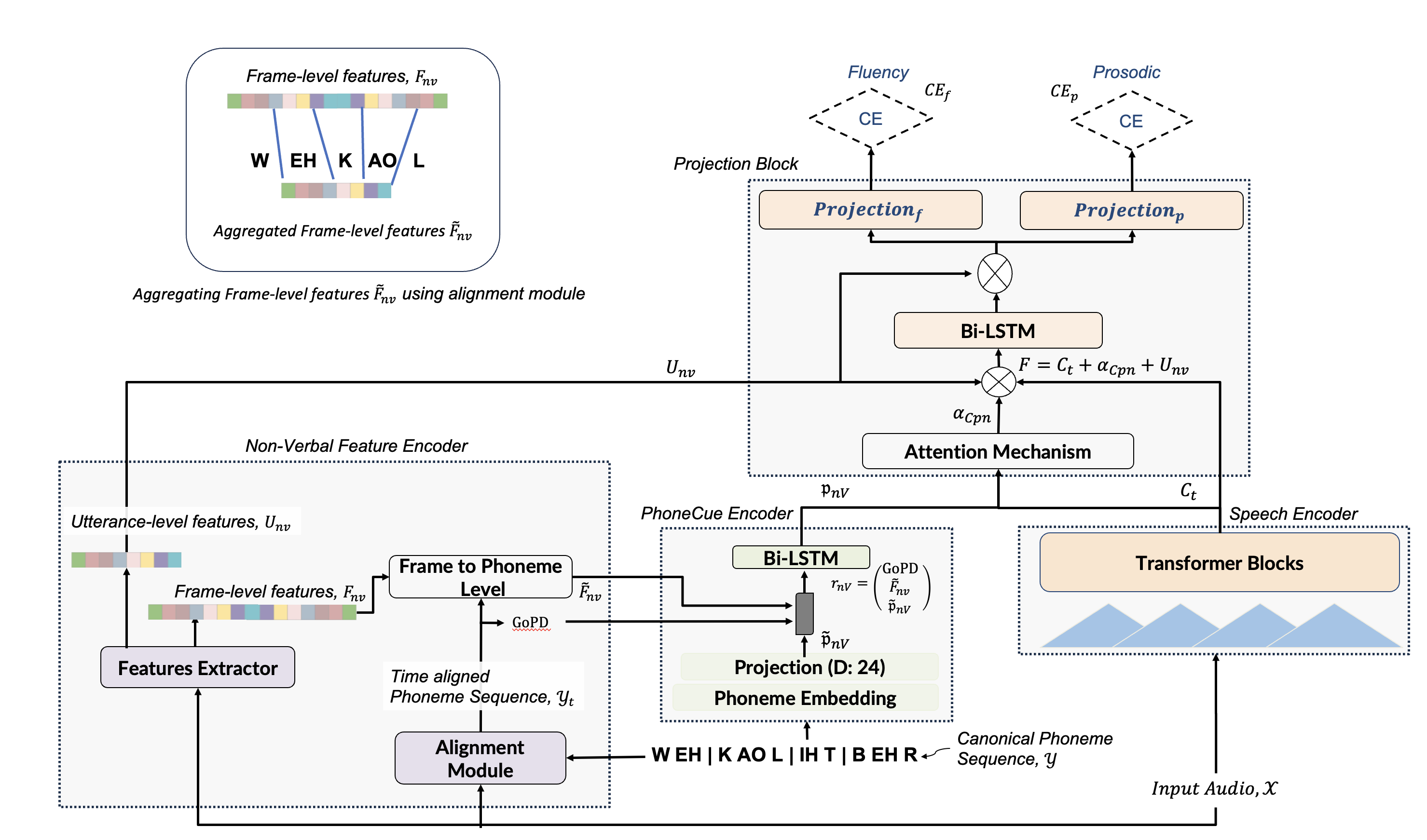}
\caption{Overview of proposed IntraVerbalPA.}
\label{fig:model}
\vspace{-0.2cm}
\end{figure*}

However, majority of the studies often neglect the rich layer of information hidden within the non-verbal cues. 
For automated PA, integrating an additional layer of non-verbal cues -- pitch, intonation, voice quality, etc., can greatly enhance the system's ability to evaluate oral proficiency, bringing in the human perception factor in the equation. 

Therefore, we introduce an novel PA framework --

\noindent\textbf{IntraVerbalPA}. The framework is jointly trained to score the `fluency' and `prosodic' aspects in multi-task setup. IntraVerbalPA leverages both latent speech and phoneme embedding, while complementing them with handcrafted frame- and utterance-level non-verbal paralinguistic cues. We explored both conventional and unconventional non-verbal cues -- duration, pitch, jitter, intonation, voice and unvoiced segment features among others, for modeling PA scores. Furthermore, we introduce a novel metric to integrate the temporal information in the model. 
The framework can seamlessly integrate any new features for both frame-level and utterance-level, hence offering opportunities to the researchers to investigate any new feature set. 
Our contributions are as follows: 
\textit{(a)} Integration of spoken non-verbal cues in the PA system;
\textit{(b)} Inclusion of frame-level low-level descriptor in IntraVerbalPA;
\textit{(c)} Utilization of utterance-level non-verbal functionals in IntraVerbalPA;
\textit{(d)} Introduction of a new metric -- Goodness of phonemic-duration, conditioned on the duration distribution of native English speakers.



\section{Proposed  Framework}
\label{sec:method}
Figure \ref{fig:model} shows our proposed IntraVerbalPA framework, designed to train an efficient end-to-end pronunciation assessment model using different sources of information from the input signal. The IntraVerbalPA model comprised of $4$ modules, \textit{Speech Encoder}, \textit{PhoneCue Encoder}, \textit{Non-Verbal Features Encoder}, and a \textit{Projection Block}. 

\paragraph*{Framework Overview} Given an input raw signal $\mathcal{X}$, of $n$ samples, we first extract contextualized acoustic representations, $C_t$ (of dimension, $D: 1024$), from the \textbf{Speech Encoder}. Simultaneously, $\mathcal{X}$ is also passed through the \textbf{Non-Verbal Features Encoder} to obtain non-verbal phoneme-level ($\mathbf{\Tilde{F}_{nv}}$), utterance-level $\mathbf{U_{nv}}$ feature along with \textit{duration}, $GoPD$ representation. 
We then pass $\mathbf{\Tilde{F}_{nv}}$ and $GoPD$ to the \textbf{PhoneCue Encoder}. The resultant output, $\mathbf{\Tilde{p}_{nv}}$, along with $C_t$, and $\mathbf{U_{nv}}$ are then pass to the \textbf{Projection Block} for predicting Fluency and Prosodic scores.

\subsection{Speech Encoder Module}
The wav2vec2-large\footnote{https://huggingface.co/facebook/wav2vec2-large-robust} \cite{conneau2020unsupervised} model is a pre-trained wav2vec2.0 \cite{baevski2020wav2vec}. It follows the same architecture as the wav2vec2.0 model. The encoder network consists of blocks of temporal convolution layers with $512$ channels, and the convolutions in each block have strides and kernel sizes that compress about $25$ms of $16$kHz audio every $20$ms. The context network consists of $24$ blocks with model dimension $1024$, inner dimension $4096$, and $16$ attention heads. 

\subsection{Non-Verbal Features Encoder}
Inside the non-verbal feature encoder, using the input $\mathcal{X}$, we first extract low level descriptors in frame-level ($\mathbf{F_{nv}}$) and apply statistical functional to create utterance-level ($\mathbf{U_{nv}}$.) representation using OpenSmile.\footnote{\url{https://audeering.github.io/opensmile-python/}}
We then align the input $\mathcal{X}$ with the canonical phoneme sequence $\mathcal{Y}$ using the \textit{Alignment Module} to convert frame-level non-verbal $\mathbf{F_{nv}}$ representation to phoneme-level ($\mathbf{\Tilde{F}_{nv}}$) representation. Moreover, we also use the phoneme-level alignments to calculate the \textit{duration representation}, $GoPD$. 

\subsubsection{Alignment Module}
\label{sec:aligner}
To align the canonical sequence with the audio, we opt for wav2vec2.0 trained for frame-level classification \footnote{https://huggingface.co/charsiu/en\_w2v2\_fc\_10ms} \cite{zhu2022charsiu}. The phone alignment is acquired through  forced alignment performed using the Dynamic Time Warping (DTW)
algorithm based on models output probability matrix and the given canonical phone transcription.

\subsubsection{Goodness of phonemic-duration (GoPD)}
We introduce a metric named Goodness of phonemic-duration (GoPD), inspired by the goodness of pronunciation (GoP) metric \cite{GOP_WITT}, which is defined for a given observation $\mathbf{O}$ and a phone $\mathbf{p}$ as follows:
\begin{equation}
    GOP(p)=P(p|O) = \frac{p(O|p) \ P(p)}{\sum_{q} \ p(O|q)\ P(q)} 
\end{equation}
First, we extracted phoneme duration from native English (subset of TIMIT \cite{garofolo1993timit}) data using the alignment module (in Section \ref{sec:aligner}). We then construct Gaussian distributions specific to each phoneme $\mathbf{p}$ denoted as $\mathbf{D_p}$ to later use it in the IntraVerbalPA framework. Within the framework, using the pre-extracted distribution, we compute the GoPD as follows:
\begin{equation}
    GoPD(d_{t}) = log(P_{D_{p_{t}}}(d_{t}))
\end{equation}
\noindent for a given duration $d_{t}$ corresponding to a L2-phoneme $p_{t}$.

\begin{figure}[htbp]
\centering
\includegraphics[width=0.5\textwidth]{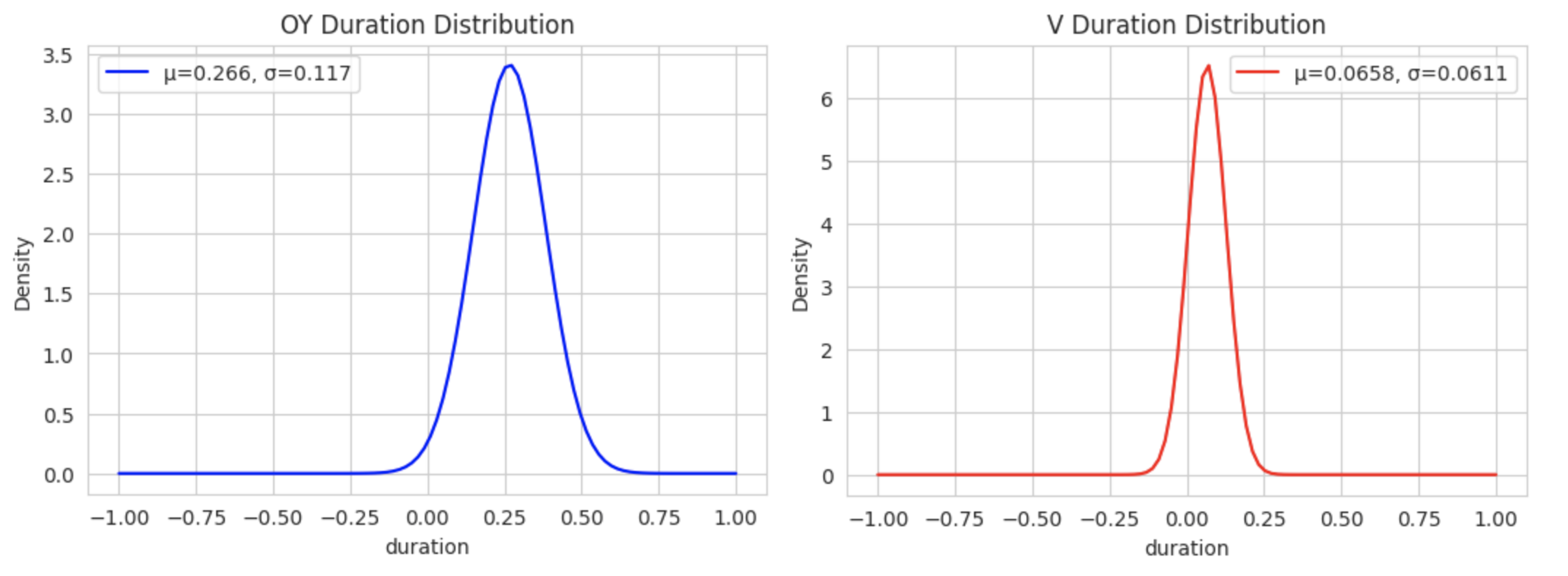}
\vspace{-0.7cm}
\caption{OY vs V duration distribution (\textit{ms}).}
\vspace{-0.6cm}
\label{fig:distribution}
\end{figure}

In Figure \ref{fig:distribution}, we present an illustration featuring two phonemes duration distributions, 'V' and 'OY'. Notably, 'OY' exhibits a relatively higher mean duration compared to 'V', which aligns with our expectations since 'V' represents a vowel sound and 'OY' a consonant sound. However, it's worth noting that 'V' displays a smaller standard deviation. This characteristic makes 'V' more sensitive to long duration, potentially signaling  elongation.


\subsubsection{Frame-level features}
As shown in Table \ref{tab:frame}, we used Loudness, AlphaRatio, Pitch, and JitterLocal extracted at frame-level using openSMILE\cite{eyben2010opensmile}. The Energy is an important feature of speech detection, the energy distribution may be related to the intonation property, we presented Energy by Loudness, and AlphaRatio. Pitch provides acoustic cues for a speaker’s intonation, confidence and expressiveness, we present that using logarithmic F0, and JitterLocal.

\begin{table}
  \centering
  \renewcommand{\arraystretch}{1} 
  \scalebox{0.7}{
  \begin{tabular}{p{2cm}p{5cm}p{3.5cm}} 
    \hline
    \textbf{Frame-level features} & \textbf{Description} & \textbf{Relevance}\\
    \hline
    \hline
    Loudness &  Estimate of perceived signal intensity from an auditory spectrum & Intonation\\
    \hline
    AlphaRatio & Ratio of the summed energy from 50-1000 Hz and 1-5 kHz. Represents the high-frequency content and the spectral balance. & Intonation\\
    \hline
    Pitch & logarithmic F0 on a semitone frequency scale & Intonation, Confidence and Expressiveness \\
    \hline
    JitterLocal & deviations in individual consecutive F0 period lengths & Intonation, Confidence and Expressiveness \\
    \hline
  \end{tabular}}
  \caption{Selected Frame-level features and their relevance}
  \label{tab:frame}
\end{table}

\subsubsection{Utterance-level features}
For utterance-level features, we selected Pitch features (mean, std, percentiles, mean rising slope, std rising slope, mean falling slope, and std falling slope) that provides information speaker’s intonation, and Voiced segment features (mean, std), Unvoiced segment features (mean, std) that provides information about disfluencies.

\subsection{PhoneCue Encoder Module}
The PhoneCue Encoder takes as input a sequence $Y = {y_1, y_2, \ldots, y_m}$ representing parsed canonical phoneme sequence, then to an embedding layer with dimension $D: 41$. These embedding are projected using a feedforward operation (with dimension $D: 24$), resulting in the intermediate feature vector $\mathbf{\Tilde{p}_{nv}}$.
Subsequently, we vertically concatenate this intermediate feature vector $\mathbf{\Tilde{p}{nv}}$ with other relevant components, including GoPD and $\mathbf{\Tilde{F}{nv}}$:
\begin{equation}
    \mathbf{r_{nv}} = \begin{bmatrix}
GoPD \\ 
 \mathbf{\Tilde{F}_{nv}} \\ 
\mathbf{\Tilde{p}_{nv}}
\end{bmatrix}
\end{equation}
Finally, the $\mathbf{r_{nv}}$ is processed through a Bi-LSTM with dimension ($D: 512$), resulting in the feature representation $\mathbf{p_{nv}}$ ($D: 1024$) capturing the non-verbal and phonetic cues present in the utterance.


\subsection{Projection Block}
The $\mathbf{p_{nv}}$ and the contextualized acoustic representations $C_t$ are then passed to a attention layer that takes $\mathbf{p_{nv}}$ as key and query, and $C_t$ as value, resulting in the final feature representation $\mathbf{\alpha_{Cnv}}$ ($D: 1024$)

\begin{equation}
    \mathbf{\alpha_{Cnv}} = Attention(K=\mathbf{p_{nv}}, Q=\mathbf{p_{nv}}, V=C_t)
\end{equation}
The embeddings $C_t$ and $\alpha_{Cp}$ ($D: 1024$) are then concatenated with utterane-level features $\mathbf{U_{nv}}$, resulting in: 
\begin{equation}
    F = C_t + \alpha_{Cpn} + \mathbf{U_{nv}}
\end{equation}

The resulting $F$ is then parsed to Bi-LSTM ($D: 512$), and get concatenated with the residual utterance-level features $\mathbf{U_{nv}}$ giving: $\Tilde{F}= BiLSTM(F) + \mathbf{U_{nv}}$. Following, $\Tilde{F}$ is then passed to two separate projection layers $Projection_{f}$, $ Projection_{p}$ of $D: 11$\footnote{11: the dimension of scores ranging from 0-10}, for respective Fluency and Prosodic score classification.

\section{Experimental Setup}
\label{sec:exp}
\subsection{Datasets}
For the study, we used the widely used Speechocean762 \cite{speechocean762}  an extensive dataset specifically designed for pronunciation assessment. The dataset comprises a total of 5,000 English utterances obtained from 250 non-native speakers. Each utterance in the dataset is associated with five aspect scores at the utterance level, namely accuracy, fluency, completeness, prosody, and a total score ranging from 0 to 10. 
To ensure reliability, each of these scores is annotated by five expert evaluators. In our study, we are interested in fluency and prosodic sentence level scores. $10\%$ of the training data is used as the validation set for the model and early stopping detection.


\subsection{Model Training and Parameters}
The models are optimized using Adam optimiser \cite{kingma2017adam} for 50 epochs with early stopping criterion ($=2$). The initial learning rate is set to $1 \times 10^{-4}$, with a batch size of $32$. For all setups, the features extraction layers remain frozen and only transformer layers are finetuned. The total loss denoted as $loss_{total}$ criterion is the combination of CE loss on fluency scores $CE_f$, and CE loss on prosodic scores $CE_p$ defined as:
\begin{equation}
\small
    loss_{total} = 0.5 \times CE_f + 0.5 \times CE_p
\end{equation}

\subsection{Evaluation}
We reported Pearson Correlation Coefficient (PCC) 
between model prediction and the average score of teachers provided in the dataset.
To overcome any randomness in the reported results, we repeated each experiment 3 times and reported average PCC. 

\section{Results and Discussion}
\label{sec:result}
Table \ref{tab:main_rslt} demonstrate the efficacy of the proposed IntraVerbalPA in modeling the Fluency and Prosodic scores. 

\noindent\textbf{Baselines:} The proposed model significantly outperform the traditional approach of modeling via fine-tuning a pre-trained model to the task, with and without encoded canonical phoneme embedding (Table \ref{tab:main_rslt}: Baseline). 


\noindent\textbf{Comparision to Prior Studies:} In comparison to contemporary models (see Table \ref{tab:main_rslt}: Contemporary), the IntraVerbalPA performs comparably with the MultiPA \cite{chen2023multipa}, and Joint-CAPT \cite{ryujoint}.
Notably, while MultiPA and Joint-CAPT operate in a multi-task setup context using either additional pre-trained features embeddings such RoBERTA, using external L2-Artic \cite{L2-ARCTIC} and TIMIT datasets, our IntraVerbalPA focuses on monolingual settings and leverages a limited L2-English Speechocean762 dataset. 

Another model, 3M \cite{3M}, integrates embeddings from three distinct self-supervised learning models —- WavLM, HuBERT, and Wav2Vec2.0 (Large), along with phoneme embeddings and additional features like transformer-based GOP, duration, and energy. Despite its more comprehensive feature set and larger number of parameters, IntraVerbalPA achieves a close results using a scaled-down features.

\noindent\textbf{Component Ablations:} The enhanced performance of IntraVerbalPA reflects the potential of non-verbal cues exploited in this work. To understand the importance of the three main proposed components: (a) frame-level integration of low-level non-verbal descriptor $\mathbf{\Tilde{F}_{nv}}$, (b) utterance level non-verbal features $\mathbf{U_{nv}}$, and (c) the Goodness of phonemic-duration $GoPD$, we performed ablation tests. The PCC reported on Table \ref{tab:main_rslt}:AblationComponents, reconfirmed the importance of integration fine-grained frame-level non-verbal cues. Removing the $\mathbf{\Tilde{F}_{nv}}$, drops the PCC by $\approx 10\%$ for both fluency and prosodic scores. We observed the duration representation, $GoPD$, also plays a key role in the model performance, specially more for prosodic information than the fluency. This aligns with previous findings showing the importance of modeling duration for PA systems. When the $\mathbf{U_{nv}}$ is removed, the performance of the IntraVerbalPA is less affected. However, the drop of absolute $1.12\%$ and $2.16\%$ in prosodic and fluency scores respectively reflects the added knowledge from $\mathbf{U_{nv}}$.

\noindent\textbf{Age Group Ablations:} PCC reported on Table \ref{tab:main_rslt}:AblationAge suggest that IntraVerbalPA performs exceptionally well for adult speech. For children's speech, the model shows satisfactory performance, although we lack prior comparative data for a detailed analysis.


\begin{table}[]
\centering
\scalebox{0.9}{
\begin{tabular}{l|cc}
\toprule
 \multicolumn{1}{c|}{\textit{PCC}} & \textbf{Prosodic} & \textbf{Fluency} \\
\midrule\midrule
\multicolumn{3}{c}{\textbf{Contemporary and Proposed Works}} \\                                                                   
\midrule\midrule
Raw Speech ($C_t$) \cite{ryujoint} & 65.00\% & 65.20\% \\
Wav2vec-large \cite{kim2022automatic} & 72.00\% & 72.00\%\\
GOPT \cite{JIM} & 76.00\% & 75.30\% \\
Joint-CAPT-L1 \cite{ryujoint} & 77.30\% & 77.50\% \\
Hubert-large-finetuned \cite{kim2022automatic} & 77.00\% & 78.00\%\\
MultiPA [Multi-Task PA] \cite{chen2023multipa} & \textit{78.7}\% & \textit{79.7}\% \\
3M \cite{3M} & \textit{82.70}\% & \textit{82.8}\% \\ \midrule\midrule
\rowcolor{blue!10}
IntraVerbalPA (Proposed) & \textbf{78.35}\% & \textbf{78.51}\% \\
\midrule
\midrule
\multicolumn{3}{c}{\textbf{Baseline Results}} \\                                                                   
\midrule\midrule
SSL:Wav2vec-large, ($Ct$) & 72.04\% & 72.00\% \\
$Ct$ + Phoneme Embedding & 70.4\% & 70.94\% \\
\midrule\midrule
\multicolumn{3}{c}{\textbf{Ablation: Components}} \\                                                                   
\midrule\midrule
\rowcolor{yellow!10}
\hspace{1cm} -- GoPD (duration) & 70.14\% & 71.60\% \\
\rowcolor{yellow!20}
\hspace{1cm} -- $\mathbf{\Tilde{F}_{nv}}$ (frame-level) & 69.26\% & 69.22\% \\
\hspace{1cm} -- $\mathbf{U_{nv}}$ (utt-level) & 77.23\% & 76.35\% \\
\midrule\midrule
\multicolumn{3}{c}{\textbf{Ablation: Age Groups}} \\
\midrule
IntraVerbalPA (Adult) & \textbf{85.31}\% & \textbf{85.47}\% \\
IntraVerbalPA (Children) & \textbf{57.95}\% & \textbf{59.80}\% \\
\midrule\midrule
\bottomrule
\end{tabular}}
\caption{Reported Pearson correlation coefficient, PCC, for the prior and contemporary works; proposed IntraVerbalPA, along with baselines and ablation of crucial components of the network. GoPD: Goodness of phonemic-duration, $\mathbf{\Tilde{F}_{nv}}$ converted phoneme-level representation of non-verbal cues, $\mathbf{U_{nv}}$: utterance-level non-verbal cues representation. }
\label{tab:main_rslt}
\end{table}

\section{Conclusion}
We introduce the IntraVerbalPA framework, enriched with both fine-grained and abstract non-verbal cues along with the conventional speech and phoneme representation for modeling pronunciation assessment system. Moreover, we propose a new metric to effectively model duration distribution within the framework.
Our reported results validate the importance of individual components of the framework, and demonstrate the efficacy of the IntraVerbalPA. The framework 
is designed to seamlessly integrate both frame-level and utterance-level information, thereby offering researchers ample opportunities to explore additional features for further investigation.

\bibliographystyle{IEEEbib}
\bibliography{qvoice}

\begin{thebibliography}{10}

\bibitem{eskenazi2009overview}
Maxine Eskenazi,
\newblock ``An overview of spoken language technology for education,''
\newblock {\em Speech Communication}, vol. 51, no. 10, pp. 832--844, 2009.

\bibitem{litman2018speech}
Diane Litman, Helmer Strik, and Gad~S Lim,
\newblock ``Speech technologies and the assessment of second language speaking:
  Approaches, challenges, and opportunities,''
\newblock {\em Language Assessment Quarterly}, vol. 15, no. 3, pp. 294--309,
  2018.

\bibitem{levy2013call}
Mike Levy and Glenn Stockwell,
\newblock {\em CALL dimensions: Options and issues in computer-assisted
  language learning},
\newblock Routledge, 2013.

\bibitem{lin20f_interspeech}
Binghuai Lin, Liyuan Wang, Xiaoli Feng, and Jinsong Zhang,
\newblock ``{Automatic Scoring at Multi-Granularity for L2 Pronunciation},''
\newblock in {\em Proc. Interspeech 2020}, 2020, pp. 3022--3026.

\bibitem{JIM}
Yuan Gong, Ziyi Chen, Iek-Heng Chu, Peng Chang, and James Glass,
\newblock ``Transformer-based multi-aspect multi-granularity non-native english
  speaker pronunciation assessment,''
\newblock in {\em ICASSP 2022 - 2022 IEEE International Conference on
  Acoustics, Speech and Signal Processing (ICASSP)}, 2022, pp. 7262--7266.

\bibitem{hu15_slate}
Wenping Hu, Yao Qian, and Frank~K. Soong,
\newblock ``{An improved DNN-based approach to mispronunciation detection and
  diagnosis of L2 learners’ speech},''
\newblock in {\em Proc. Speech and Language Technology in Education (SLaTE
  2015)}, 2015, pp. 71--76.

\bibitem{zhang2021multilingual}
Huayun Zhang, Ke~Shi, and Nancy~F Chen,
\newblock ``Multilingual speech evaluation: case studies on english, malay and
  tamil,''
\newblock {\em arXiv preprint arXiv:2107.03675}, 2021.

\bibitem{3M}
Fu-An Chao, Tien-Hong Lo, Tzu-I Wu, Yao-Ting Sung, and Berlin Chen,
\newblock ``3m: An effective multi-view, multi-granularity, and multi-aspect
  modeling approach to english pronunciation assessment,''
\newblock in {\em 2022 Asia-Pacific Signal and Information Processing
  Association Annual Summit and Conference (APSIPA ASC)}, 2022, pp. 575--582.

\bibitem{chen2023multipa}
Yu-Wen Chen, Zhou Yu, and Julia Hirschberg,
\newblock ``Multipa: a multi-task speech pronunciation assessment system for a
  closed and open response scenario,'' 2023.

\bibitem{kim2022automatic}
Eesung Kim, Jae-Jin Jeon, Hyeji Seo, and Hoon Kim,
\newblock ``Automatic pronunciation assessment using self-supervised speech
  representation learning,'' 2022.

\bibitem{10022486}
Binghuai Lin and Liyuan Wang,
\newblock ``Exploiting information from native data for non-native automatic
  pronunciation assessment,''
\newblock in {\em 2022 IEEE Spoken Language Technology Workshop (SLT)}, 2023,
  pp. 708--714.

\bibitem{yang2022improving}
Mu~Yang, Kevin Hirschi, Stephen~D Looney, Okim Kang, and John~HL Hansen,
\newblock ``Improving mispronunciation detection with wav2vec2-based momentum
  pseudo-labeling for accentedness and intelligibility assessment,''
\newblock {\em arXiv preprint arXiv:2203.15937}, 2022.

\bibitem{conneau2020unsupervised}
Alexis Conneau, Alexei Baevski, Ronan Collobert, Abdelrahman Mohamed, and
  Michael Auli,
\newblock ``Unsupervised cross-lingual representation learning for speech
  recognition,'' 2020.

\bibitem{baevski2020wav2vec}
Alexei Baevski, Yuhao Zhou, Abdelrahman Mohamed, and Michael Auli,
\newblock ``wav2vec 2.0: A framework for self-supervised learning of speech
  representations,''
\newblock {\em Advances in neural information processing systems}, vol. 33, pp.
  12449--12460, 2020.

\bibitem{zhu2022charsiu}
Jian Zhu, Cong Zhang, and David Jurgens,
\newblock ``Phone-to-audio alignment without text: A semi-supervised
  approach,''
\newblock {\em IEEE International Conference on Acoustics, Speech and Signal
  Processing (ICASSP)}, 2022.

\bibitem{GOP_WITT}
Silke~M Witt and Steve~J Young,
\newblock ``Phone-level pronunciation scoring and assessment for interactive
  language learning,''
\newblock {\em Speech communication}, vol. 30, no. 2-3, pp. 95--108, 2000.

\bibitem{garofolo1993timit}
John~S Garofolo,
\newblock ``Timit acoustic phonetic continuous speech corpus,''
\newblock {\em Linguistic Data Consortium, 1993}, 1993.

\bibitem{eyben2010opensmile}
Florian Eyben, Martin W{\"o}llmer, and Bj{\"o}rn Schuller,
\newblock ``Opensmile: the munich versatile and fast open-source audio feature
  extractor,''
\newblock in {\em Proceedings of the 18th ACM international conference on
  Multimedia}, 2010, pp. 1459--1462.

\bibitem{kingma2017adam}
Diederik~P. Kingma and Jimmy Ba,
\newblock ``Adam: A method for stochastic optimization,'' 2017.

\bibitem{ryujoint}
Hyungshin Ryu, Sunhee Kim, and Minhwa Chung,
\newblock ``A joint model for pronunciation assessment and mispronunciation
  detection and diagnosis with multi-task learning,''
\newblock .

\bibitem{L2-ARCTIC}
Guanlong Zhao, Sinem Sonsaat, Alif Silpachai, Ivana Lucic, Evgeny
  Chukharev-Hudilainen, John Levis, and Ricardo Gutierrez-Osuna,
\newblock ``L2-arctic: A non-native english speech corpus.,''
\newblock in {\em Interspeech}, 2018, pp. 2783--2787.

\end{thebibliography}

\end{document}